\documentclass[preprint,12pt]{elsarticle}

\usepackage{makecell}
\usepackage{amsmath}
\usepackage{amssymb}
\usepackage{hyperref}
\hypersetup{pdfauthor={Hossein Amirkhani}}
\usepackage{xpatch}
\usepackage{xcolor}
\usepackage[utf8]{inputenc}
\usepackage[arabic,farsi,english]{babel}
\usepackage{multirow}
\usepackage{array}
\usepackage{subcaption}

\newcolumntype{P}[1]{>{\centering\arraybackslash}p{#1}}
\setcitestyle{square}
\makeatletter

\begin{document}

\begin{frontmatter}



\title{FarsTail: A Persian Natural Language Inference Dataset}


\author{Hossein Amirkhani} \ead{amirkhani@qom.ac.ir}
\author{Mohammad AzariJafari}
\author{Zohreh Pourjafari}
\author{\\Soroush Faridan-Jahromi}
\author{Zeinab Kouhkan}
\author{Azadeh Amirak}
\address{Computer Engineering and IT Department, University of Qom, Iran}
\begin{abstract}
Natural language inference (NLI) is known as one of the central tasks in natural language processing (NLP) which encapsulates many fundamental aspects of language understanding. With the considerable achievements of data-hungry deep learning methods in NLP tasks, a great amount of effort has been devoted to develop more diverse datasets for different languages. In this paper, we present a new dataset for the NLI task in the Persian language, also known as Farsi, which is one of the dominant languages in the Middle East. This dataset, named FarsTail, includes 10,367 samples which are provided in both the Persian language as well as the indexed format to be useful for non-Persian researchers. The samples are generated from 3,539 multiple-choice questions with the least amount of annotator interventions in a way similar to the SciTail dataset. A carefully designed multi-step process is adopted to ensure the quality of the dataset. We also present the results of traditional and state-of-the-art methods on FarsTail including different embedding methods such as word2vec, fastText, ELMo, BERT, and LASER, as well as different modeling approaches such as DecompAtt, ESIM, HBMP, and ULMFiT
to provide a solid baseline for the future research. The best obtained test accuracy is 83.38\% which shows that there is a big room for improving the current methods to be useful for real-world NLP applications in different languages. We also investigate the extent to which the models exploit superficial clues, also known as dataset biases, in FarsTail, and partition the test set into \textit{easy} and \textit{hard} subsets according to the success of biased models. The dataset is available at \url{https://github.com/dml-qom/FarsTail}. 
\end{abstract}




\end{frontmatter}


\section{Introduction}
\label{sec:introduction}
Natural Language Processing (NLP) deals with the development of automatic methods for processing, analyzing, and generating human languages. It consists of a vast number of problems, ranging from low-level to high-level tasks such as named entity recognition~\citep{yadav2019survey}, sentiment analysis~\citep{keramatfar2019bibliometrics}, machine translation~\citep{yang2020survey}, and machine reading comprehension~\citep{baradaran2020survey}. One important task in NLP is Natural Language Inference (NLI) which is believed to be a stringent test for language understanding, since a system with the ability to identify the implications of natural language sentences should have a good level of language understanding~\citep{maccartney2009natural}.

The goal of NLI is to determine the inference relationship between a premise $p$ and a hypothesis $h$. It is a three-class problem, where each pair $(p,h)$ is assigned to one of these classes: \textit{entailment} if the hypothesis can be inferred from the premise, \textit{contradiction} if the hypothesis contradicts with the premise, and \textit{neutral} if none of the other conditions hold. To determine the hypothesis status, some prior knowledge is considered besides the premise. This includes the knowledge that typical speakers of that language know, such as the commonsense facts and general semantic knowledge. For example, the typical English speakers know that “USA” refers to “the United States of America”.

After substantial success of deep learning (DL) based methods in different artificial intelligence tasks, the NLP researchers also started to develop DL-based models to learn the patterns in available natural language data generated by humans~\citep{otter2020survey}.  The percentage of deep learning papers nearly doubled in a six-year period from 2012 in the major NLP conferences~\citep{young2018recent}. Since these methods need a large amount of training data to let the model learn the general pattern for the particular task without overfitting to the available data, different research groups started to gather and publish large datasets. For the NLI task, the development of Stanford NLI dataset (SNLI) caused a considerable progress in developing DL-based models for NLI task~\citep{bowman2015large}.

In DL-based NLI literature, there has been a considerable amount of researches on languages with a large amount of training data, such as English, but relatively little attention has been paid to data-poor languages. Despite some efforts in developing NLI datasets for other languages by translation or transferring knowledge obtained from learning on one language to other languages~\citep{conneau2018xnli}, presenting native datasets for other languages help develop models with more comprehensive language understanding capabilities. In addition, these datasets can be used to evaluate the proposed learning architectures and methods for a broader range of languages. 

The focus of this paper is on Persian (Farsi) language which is a pluricentric language spoken and used by around 110 million people in countries such as Iran, Afghanistan, and Tajikistan. It has had a considerable influence on its neighboring languages such as Turkic, Armenian, Georgian, and Indo-Aryan languages. Its alphabet includes 32 characters written right to left. Table~\ref{tab:farsi-difficulties} shows some features of Persian language which make its processing different from other languages.

\begin{table}[!t]
\caption{Some features of Persian language which make its processing different from other languages.}
    \label{tab:farsi-difficulties}
    \centering
    \includegraphics[width=\textwidth]{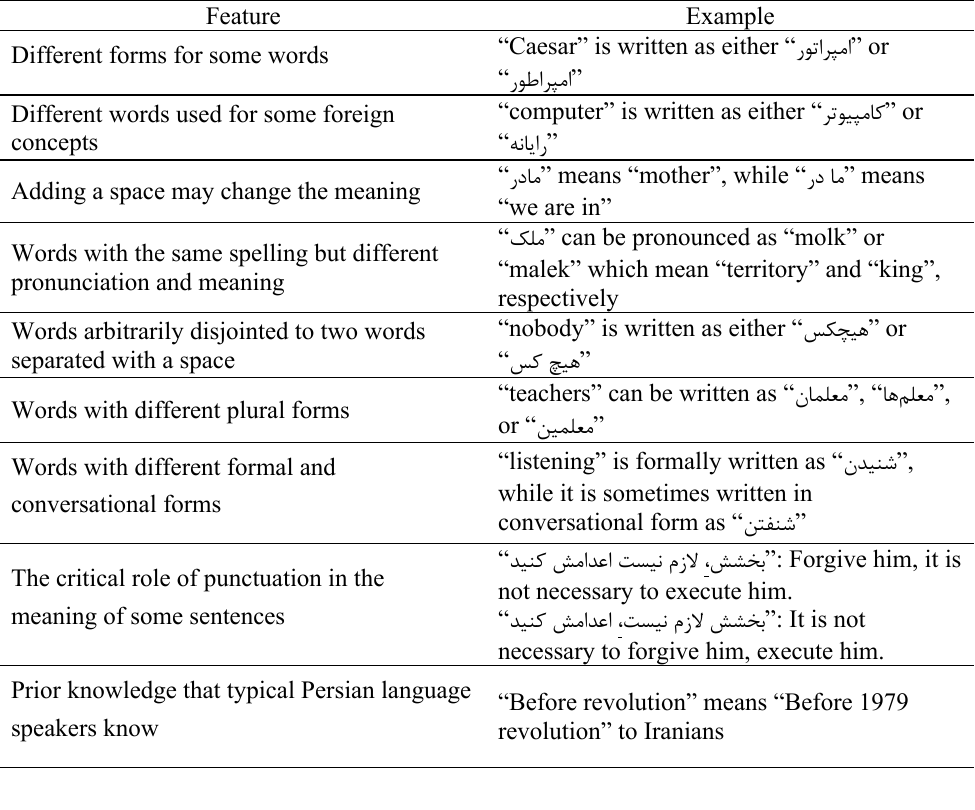}
\end{table}

In this paper, we present, to the best of our knowledge, the first relatively large-scale Persian corpus for NLI task, called FarsTail. We tried to reduce the amount of annotation interventions to provide realistic samples which are naturally occurring in real-world applications instead of task-specific synthesized examples. A protocol similar to the SciTail dataset~\citep{khot2018scitail} is followed where the sentences are either generated, with the least amount of interventions, from multiple-choice questions or selected from natural sentences that already exist independently “in the wild”. However, in contrast to SciTail which only includes the \textit{neutral} and \textit{entailment} classes, we also include \textit{contradiction} examples in the dataset.  


Each person generates three data examples from a multiple-choice question, one for each class, with the same premises but different hypotheses. The \textit{entailment} hypothesis is formed by substituting the correct answer in the question. Then, a text snippet is extracted from web that the generated hypothesis can be inferred from. The \textit{contradiction} hypothesis is formed by substituting one wrong answer in the question. Finally, the \textit{neutral} hypothesis is extracted from web such that it is similar to the question but with an unknown status based on the premise. In the next phase, each sample is relabeled by four other persons and the samples with at least 4 out of 5 agreements are preserved. The rejected samples undergo a new modification and relabeling phase. 

A total of 10,367 samples are generated from a collection of 3,539 multiple-choice questions. The train, validation, and test portions include 7,266, 1,537, and 1,564 instances, respectively. We ensure that the instances with the same premises are in the same set. The developed dataset can also be used in other tasks such as question answering, summarization, semantic search, and machine translation. The developed dataset (as raw texts for Persian researchers and indexed data for non-Persian researchers) has been released for non-commercial usages. 

We evaluate different traditional and state-of-the-art methods on FarsTail, including different embedding methods such as word2vec~\citep{mikolov2013distributed}, fastText~\citep{bojanowski2017enriching}, ELMo~\citep{peters2018deep}, BERT~\citep{devlin2018bert}, and LASER~\citep{artetxe2019massively}, as well as different modeling methods such as DecompAtt~\citep{parikh2016decomposable}, ESIM~\citep{chen2016enhanced}, HBMP~\citep{talman2019sentence}, and ULMFiT~\citep{howard2018universal}. 
The best obtained accuracy on test set is 83.38\% which shows that there are many rooms to improve the models trained on this dataset. We also investigate the superficial clues, also known as dataset biases, available in FarsTail to obtain a more realistic view of the performance of the models.


Concurrent to this work, ParsiNLU~\citep{khashabi2020parsinlu} is developed which is a suite of Persian datasets for different tasks, including an NLI set with 2,700 instances. Around half of the instances are written by native speakers and the remaining instances are translated from the MNLI dataset~\citep{williams2017broad}. FarsTail is superior to this dataset in three aspects: It has around 4 times more instances; it just includes first-hand native sentences without translation clues; and task-specific human-generated texts are kept as low as possible to provide instances which are naturally occurring in real-world applications.

The rest of this paper is organized as follows. In Section~\ref{sec:literature}, the available English and non-English NLI datasets are reviewed. Section~\ref{sec:farstail} presents the FarsTail development process as well as its statistics. The experimental results are presented in Section~\ref{sec:experiments}, and the paper concludes in the last section. 

\section{Related work}
\label{sec:literature}
In this section, we review some available English and non-English NLI datasets. 

\subsection{English NLI datasets}
\begin{itemize}
    \item SICK~\citep{marelli2014semeval}: As one of the first attempts to introduce relatively large-scale datasets for NLI task, this dataset was introduced as a task in SemEval-2014. It consists of about 10k English sentence pairs annotated for two different tasks, relatedness in meaning and entailment. The original sentence pairs are randomly selected from 8k ImageFlickr dataset and the SemEval 2012 STS MSR-Video Description dataset. Some rule-based syntactic and lexical transformations are applied to each sentence to obtain sentences with similar, contradictory, and different meanings. Its partly automated construction introduced some spurious patterns into the data~\citep{bowman2015large}. 
    \item SNLI~\citep{bowman2015large}: The Stanford NLI dataset has been developed to alleviate the lack of large-scale annotated data for the NLI problem. It includes 570k labeled instances (550k training, 10k validation, and 10k test examples) gathered using the Amazon Mechanical Turk. An image caption was presented to each turker as the premise and they were asked to generate three sentences as hypothesis, one for each class (entailment, contradiction, and neutral). In the relabeling phase, if at least three out of four new labelers agreed with the main label, this instance was kept in the dataset. This dataset played a considerable role in developing and enhancing deep learning-based NLI systems.  
    \item MultiNLI~\citep{williams2017broad}: Compared to SNLI, MultiNLI covers 10 different genres of spoken and written text. With 433k instances, its scale is comparable to SNLI. The test set consists of two parts: matched set which includes the same genres in the training set and mismatched set which includes genres not available in the training set. This allows for cross-genre generalization evaluation. 
    \item MedNLI~\citep{romanov2018lessons}: This dataset was generated by the same approach as SNLI, adjusted for the clinical domain. The MIMIC-III v1.3~\citep{johnson2016mimic}, with de-identified records of 38,597 patients, was used as the premise source. The hypothesis sentences were generated by clinicians. Four clinicians worked on a total of 4,683 premises over a period of six weeks, which resulted in 14,049 unique sentence pairs. 
    \item SciTail~\citep{khot2018scitail}: This is the first NLI dataset which is collected using the available texts without authoring the sentences. This makes the dataset more realistic, since it consists of natural texts instead of task-specific synthesized sentences. SciTail is the most similar dataset to the dataset presented in this paper. The hypotheses were created from science questions and their corresponding answers, and premises were gathered from the relevant web sentences. It contains 1,834 questions with 10,101 entailment instances and 16,925 neutral ones. This dataset does not contain the contradiction label. 
    \item QA-NLI~\citep{demszky2018transforming}: This dataset is similar to SciTail, except that it was fully automatically constructed. The authors proposed a method to derive NLI datasets from the question answering datasets. This was done by introducing the QA2D task to derive a declarative sentence from a question-answer pair. The generated sentence $(D)$ along with the corresponding passage $(P)$ forms an NLI example as $(P,D)$. For the correct, incorrect, and unknown answers, the pairs were labeled as entailment, contradiction, and neutral, respectively. Note that incorrect answers are available in QA datasets with multiple answers, and unknowns are also available in some datasets such as SQuAD 2.0~\citep{rajpurkar2018know}. 
\end{itemize}

\subsection{Non-English NLI datasets}
\begin{itemize}
    \item Evalita~\citep{bos2009textual}: Constructed on the basis of Wikipedia revision histories, this dataset includes 800 short Italian sentence pairs.
    \item ArbTEDS~\citep{alabbas2013dataset}: This is a small Arabic dataset with 600 pairs annotated as either inferable or non-inferable. A semi-automatic tool was used to extract the candidate pairs from web, using the Arabic news headlines as the hypothesis and one paragraph returned by the Google-API for this headline as the premise. The pairs were then labeled by eight annotators. 
    \item German emails~\citep{eichler2014analysis}:  Constructed from the customer emails to the support center of a multimedia software company as premises and the category descriptions as the hypotheses, this dataset includes 638 entailment and 24,143 non-entailment pairs.  The matching and non-matching categories were considered as entailment and non-entailment hypotheses, respectively. 
    \item ASSIN~\citep{fonseca2016overview}: This is a two-class dataset with the entailment and not-entailment classes including a collection of 10,000 pairs, half in Brazilian Portuguese and half in European Portuguese. 
    \item XNLI~\citep{conneau2018xnli}: This dataset was developed for evaluating the cross-lingual understanding capabilities of models. The same crowdsourcing-based procedure used for MultiNLI dataset~\citep{williams2017broad} was followed to collect and validate 750 examples from each of ten text sources resulted in a total of 7,500 examples. These examples were then translated into 14 different languages by professional translators. The total 112,500 annotated pairs are in English, French, Spanish, German, Greek, Bulgarian, Russian, Turkish, Arabic, Vietnamese, Thai, Chinese, Hindi, Swahili, and Urdu languages. Unfortunately, it does not include the Persian language. 
    \item OCNLI~\citep{hu2020ocnli}: This is the first large-scale Chinese NLI dataset which includes around 56k annotated sentence pairs. The annotations are elicited from native speakers specializing in linguistics.
    \item ParsiNLU~\citep{khashabi2020parsinlu}: This concurrent work is a suite of Persian datasets for different tasks, including an NLI set with 2,700 instances. Around half of the instances are written by native speakers and the remaining instances are translated from the MultiNLI dataset~\citep{williams2017broad}. The superiority of the FarsTail dataset over the ParsiNLU NLI set is that it includes around 4 times more instances which are first-hand native sentences without translation clues. Also, to provide texts that are naturally occurring in real-world applications, FarsTail includes the least amount of task-specific human-generated texts.
\end{itemize}

\section{FarsTail dataset}
\label{sec:farstail}
\begin{figure}[t!]
    \centering
    \includegraphics[width=\textwidth]{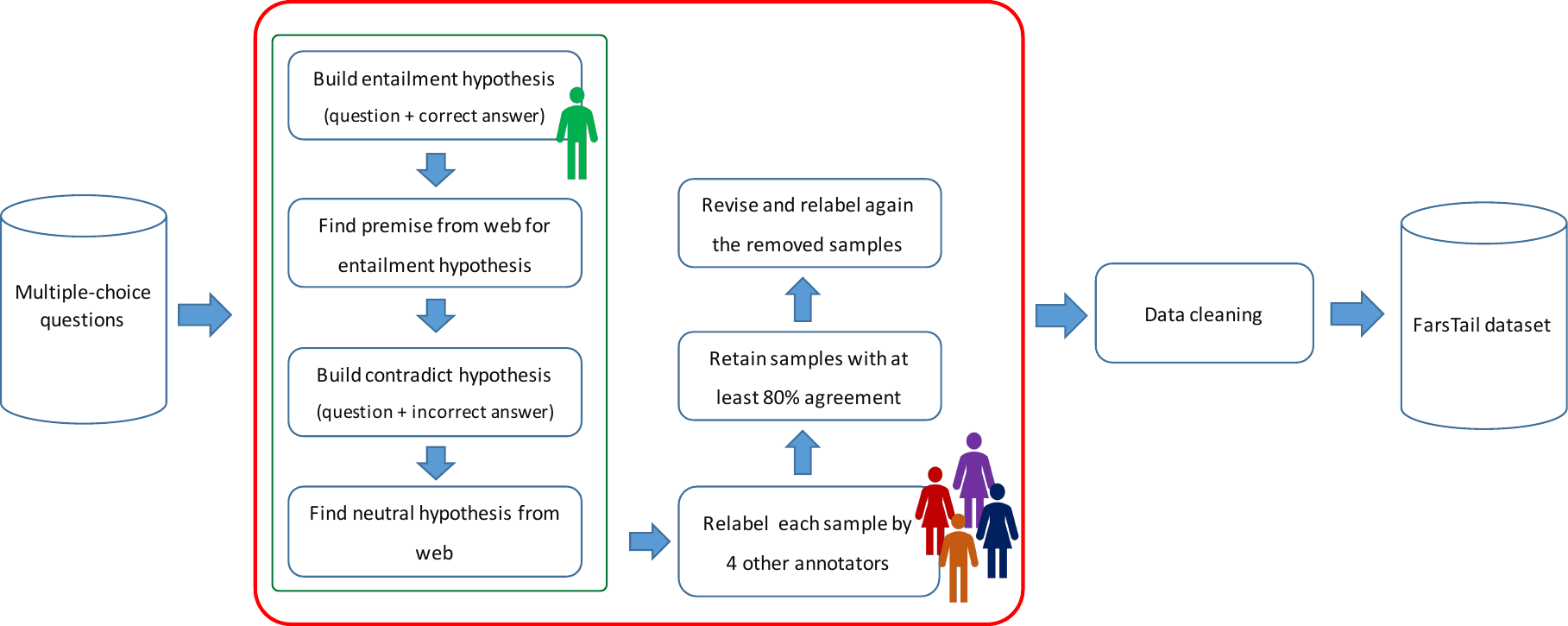}
    \caption{The FarsTail dataset development steps.}
    \label{fig:farstail-steps}
\end{figure}

In this section, we present the process of developing FarsTail dataset as well as its statistics. FarsTail has been developed with a process similar to the SciTail dataset~\citep{khot2018scitail} with some modifications. A group of five persons (called annotators herein) with a background in NLI worked under the supervision of an NLP expert to develop FarsTail. The taken steps are depicted in Fig.~\ref{fig:farstail-steps} which include generating NLI instances from multiple-choice questions, relabeling, and data cleaning. The details of these steps are given in Sections~\ref{subsec:generate}~and~\ref{subsec:relabel}, and the dataset statistics are presented in Section~\ref{subsec:statistics}. 

\subsection{Generating NLI instances from questions}
\label{subsec:generate}
\begin{figure}[t!]
    \centering
    \includegraphics[width=\textwidth]{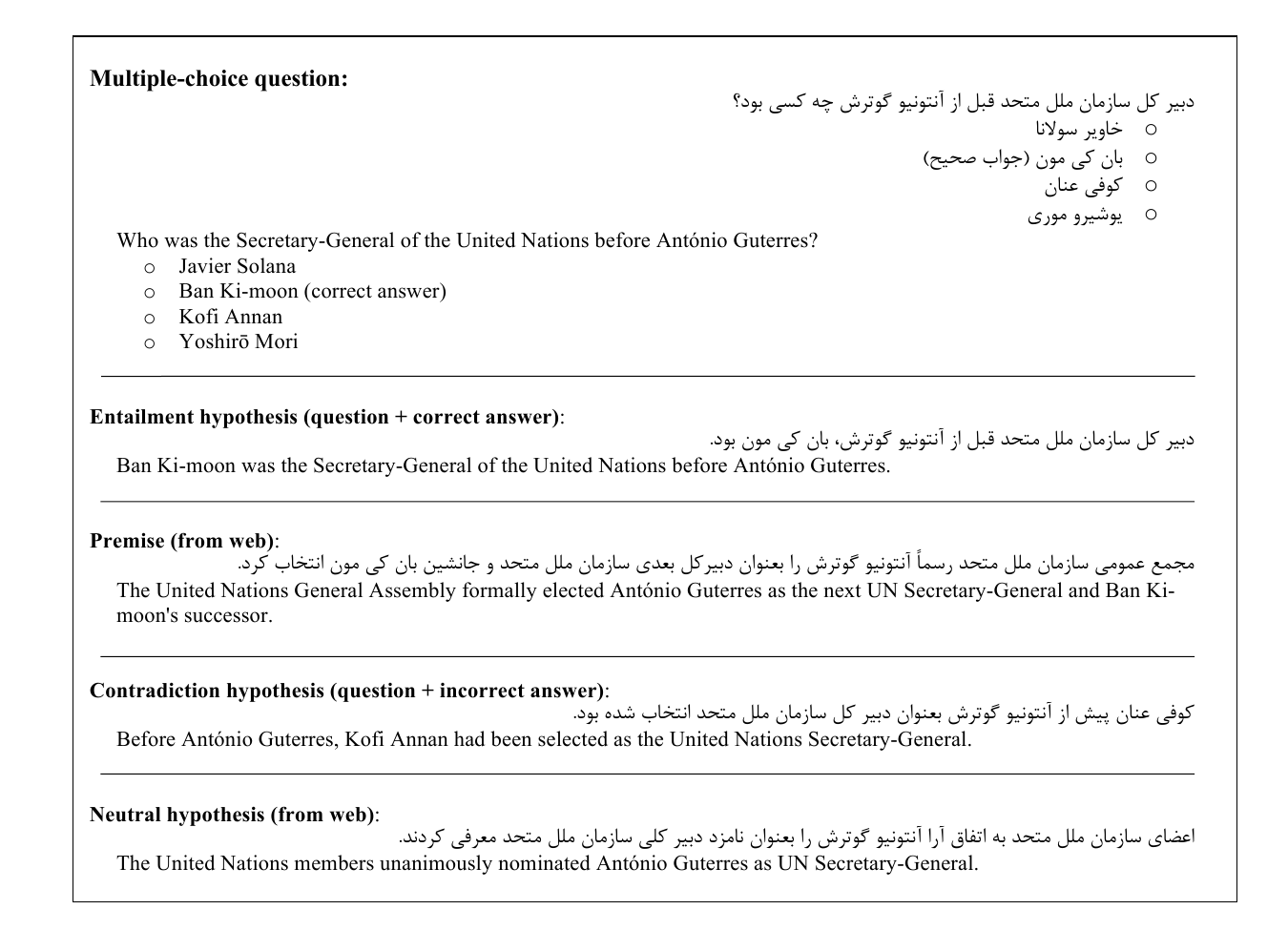}
    \caption{An example of generating NLI instances from questions in FarsTail.}
    \label{fig:generation}
\end{figure}

A collection of 3,539 multiple-choice questions was gathered from Iranian university exams in different topics including religion, history, constitution of Iran, history of literature, and Islamic revolution. For each multiple-choice question, an annotator followed the following steps to generate three different pairs, one for each class (entailment, contradiction, and neutral):
\begin{enumerate}
    \item The correct answer is inserted into the question to generate a sentence called $h_1$. 
    \item The web is searched to find a text portion $p$ where  $(p,h_1)$ has entailment relation. We use the available texts on the web instead of generating the premises to provide real-world, naturally occurring texts instead of task-specific synthesized examples. 
    \item An incorrect answer is inserted into the question to generate a sentence called $h_2$ such that $(p,h_2)$ has contradiction relation. The annotator is asked to generate $h_2$ similar to $h_1$ in length, but different in structure and words. 
    \item From the web, a related sentence $h_3$ is found with a similar length to $h_1$ and $h_2$ such that its entailment or contradiction relation cannot be inferred from $p$. The pair $(p,h_3)$ is considered as a neutral instance.
\end{enumerate}
Fig.~\ref{fig:generation} shows an example of the sample generation process in FarsTail. 

\subsection{Relabeling and data cleaning}
\label{subsec:relabel}
After the sample generation phase, each sample was relabeled by the other four annotators retaining the samples with an agreement of at least 80\% among five labelers. The samples were presented to the annotators in a random order to reduce annotation bias caused by presenting the samples with the same premise in succession. To give the rejected samples one more chance, they were revised by their original annotator and relabeled again. The samples which could not obtain a 80\% label agreement in any of these two relabeling phases were removed. Among all 10,617 samples $(3,539\times3)$, 190 samples were removed in this phase resulting in 10,427 instances.

The retained samples were investigated one more time for spelling and writing mistakes emphasizing on avoiding probable label change caused by cleaning. Finally, to reduce the unwanted repetition in the data, 60 more samples were removed including the instances generated from different questions which both their premises and hypotheses had a cosine similarity higher than 0.8. The total number of samples in the dataset is therefore 10,367. 

The instances were randomly divided into training, validation, and test sets such that the samples generated from the same question were in the same subset. In addition, to avoid information leak, the samples generated from different questions which either their premises or hypotheses had a cosine similarity higher than 0.9 were included in the same subset. The training, validation, and test sets percentages are nearly 70/15/15 with 7,266, 1,537, and 1,564 samples, respectively.

The dataset is presented in two formats, raw and indexed. The raw data includes the Persian sentences, while the indexed data is a tokenized version of sentences where each sentence is encoded as a list of word indexes (integers)\footnote{Hazm python library was used for tokenization (\url{https://github.com/sobhe/hazm})}. 

\subsection{FarsTail statistics}
\label{subsec:statistics}
The statistics of FarsTail dataset is presented in Table~\ref{tab:statistics}. To provide the possibility for comparing different subsets, there is one section for each of train, validation, and test sets. For each of these sets, beside the total statistics, the statistics for different classes are also shown separately where E, C, and N stand for \textit{entailment}, \textit{contradiction}, and \textit{neutral} classes, respectively.

\begin{table}[t!]
    \centering
    \caption{Statistics of the FarsTail dataset.}
    \label{tab:statistics}
    \begin{tabular}{c c|P{1.2cm} P{1.2cm} P{1.2cm} P{1.2cm} P{1.2cm} P{1.2cm} P{1.2cm}}
        \Xhline{2\arrayrulewidth}
         subset & class & \parbox{1.2cm}{\centering samples} & \parbox{1.2cm}{\centering prem.\\ tokens} & \parbox{1.2cm}{\centering hyp.\\ tokens} & \parbox{1.2cm}{\centering prem.\\ proc.\\ tokens} & \parbox{1.2cm}{\centering hyp.\\ proc.\\ tokens} & overlap & \parbox{1.2cm}{\centering proc.\\ overlap} \\
         \hline
         \multirow{4}{*}{Train} & E & 2,429 & 40.50 & 15.53 & 19.35 & 8.42 & 0.67 & 0.68\\
         & N & 2,448 & 40.52 & 15.62 & 19.31 & 8.26 & 0.40 & 0.30\\
         & C & 2,389 & 40.23 & 15.61 & 19.20 & 8.30 & 0.57 & 0.54\\
         & Total & 7,266 & 40.42 & 15.59 & 19.29 & 8.33 & 0.55 & 0.51\\
         \hline
         \multirow{4}{*}{Val} & E & 515 & 39.70 & 14.85 & 19.13 & 8.27 & 0.67 & 0.66\\
         & N & 523 & 39.71 & 14.95 & 19.16 & 8.06 & 0.39 & 0.29\\
         & C & 499 & 39.58 & 15.09 & 19.17 & 8.11 & 0.58 & 0.54\\
         & Total & 1,537 & 39.67 & 14.96 & 19.15 & 8.14 & 0.54 & 0.50\\
         \hline
         \multirow{4}{*}{Test} & E & 519 & 39.57 & 15.48 & 18.84 & 8.39 & 0.68 & 0.68\\
         & N & 535 & 39.23 & 16.02 & 18.73 & 8.36 & 0.38 & 0.27\\
         & C & 510 & 39.44 & 15.81 & 18.86 & 8.38 & 0.57 & 0.52\\
         & Total & 1,564 & 39.41 & 15.78 & 18.81 & 8.38 & 0.54 & 0.49\\
         \Xhline{2\arrayrulewidth}
    \end{tabular}
\end{table}

The column ``samples" of Table~\ref{tab:statistics} shows the number of samples in each subset. As mentioned in Section~\ref{subsec:relabel}, 70/15/15\% of data go to the train, validation, and test sets, respectively. It can be seen that this is a balanced dataset without any meaningful differences between the number of samples in different classes. 

The next column (premise tokens) presents the average number of tokens in the premises obtained by the Hazm python library's tokenizer. The next column (hypothesis tokens) shows the same values for hypothesis sentences. To provide a more meaningful length statistic, the next two columns (premise processed tokens and hypothesis processed tokens) report the number of unique tokens ignoring stopwords\footnote{A stoplist with 389 words was used from Hazm library.} as well as one-character tokens including punctuations. It is worth mentioning that there are a total of 20,973 tokens in FarsTail dataset where 467 tokens are stopwords or one-character tokens. 

According to these four ``tokens" columns, there is not any significant difference between the average number of tokens in train, validation, and test sets. More importantly, the average number of tokens in different classes are almost the same which shows that the length of premises and hypotheses cannot be exploited as a feature to find clues about the class of the given inputs. 

One more point to consider about the ``tokens" columns is that the premises in FarsTail are longer than the premises in SciTail dataset~\citep{khot2018scitail}. The reported average premise length for \textit{entail} and \textit{neutral} samples in SciTail training set are 10.79 and 10.28, respectively, while these numbers are 19.35 and 19.31 in FarsTail. Regarding hypotheses, the average length for \textit{entail} and \textit{neutral} samples are respectively 6.69 and 7.01 which are almost the same as FarsTail (8.42 and 8.26). These longer premises are due to the FarsTail's sample generation process where we insisted on finding exact web text portions which the hypothesis could be inferred from. Anyway, this makes FarsTail a more challenging dataset since it seeks more reasoning to connect the facts presented in longer premises. 

Finally, the last two columns show the average proportion of the hypothesis tokens that overlap with the premise. Both columns treat the sentences as a set of tokens ignoring the word repetition, but the second column also ignore the stopwords and one-character tokens. As expected, the most and the least overlap between premise and hypothesis are in the \textit{entailment} and \textit{neutral} samples, respectively. This shows that there are some superficial clues in the samples which can be exploited to estimate the relationship between two sentences without truly understanding them. In Section~\ref{subsec:bias}, we show that the mere similarity between premise and hypothesis can be used in a simple baseline model which obtains an accuracy higher than random; however, this accuracy is far from what that is obtained by more advanced deep models.

\section{Experiments}
\label{sec:experiments}
In this section, we present the results of different methods on the FarsTail dataset to provide a baseline for future researches. The evaluated models are introduced in Sections~\ref{subsec:models} and the results are presented in Section~\ref{subsec:results}. Finally, in Section~\ref{subsec:bias}, we investigate the biases available in FarsTail to provide a more realistic view of the performance of the models.

\subsection{Models}
\label{subsec:models}
We used different methods for representing the input sentences ranging from traditional TF-IDF to more recent word embedding methods such as word2vec\footnote{\url{http://vectors.nlpl.eu/repository}}~\citep{mikolov2013distributed}, fastText\footnote{\url{https://fasttext.cc/docs/en/crawl-vectors.html}}~\citep{bojanowski2017enriching}, ELMo\footnote{\url{https://github.com/HIT-SCIR/ELMoForManyLangs}}~\citep{peters2018deep}, and BERT~\citep{devlin2018bert}.
For the BERT method, we fine-tuned two pre-trained models from the Hugging Face Transformers library~\citep{wolf2019huggingface}, ParsBERT~\citep{farahani2020parsbert} and BERT-base-multilingual-cased (mBERT).

As the classifier, we exploited different methods including Support Vector Machine (SVM), Long Short-Term Memory (LSTM), and Gated Recurrent Unit (GRU) along with three models developed specially for the NLI task including DecompAtt~\citep{parikh2016decomposable}, ESIM~\citep{chen2016enhanced}, and HBMP~\citep{talman2019sentence}. 

One popular approach in learning with small labeled training datasets is to train a language model (LM) on a large unlabeled corpus and fine-tune it on the downstream task. Besides the BERT-based models which lie in this category, we tested ULMFiT~\citep{howard2018universal} with three steps: LM pre-training, LM fine-tuning, and classifier fine-tuning. In the first step, a language model was trained on a general-domain corpus. We used the Persian Wikipedia for this purpose. Then, the trained LM was fine-tuned on the target task texts without considering their labels. Finally, the pre-trained language model was augmented with additional layers which were trained on the labeled dataset of the target task. 


We also tested LASER\footnote{\url{https://github.com/facebookresearch/LASER}}~\citep{artetxe2019massively} as an embedding space which is shared between multiple languages. Since LASER provides sentence embeddings rather than word embeddings, a simple deep model was trained on the computed representations. 

The hyper-parameters were chosen based on the models' accuracy on the validation set. Most importantly, we selected the following values for the BERT models: 3 epochs of training with a learning rate of 2e-5, a batch size of 32, and a weight decay of 0.5.

\subsection{Results}
\label{subsec:results}

Table~\ref{tab:overall-results} shows the results obtained from training different models on the FarsTail training set. 
Note that the LASER and tf-idf representations were just used with the SVM classifier because they deliver sentence-level representations which cannot be used with the word-level methods like LSTM and BiGRU. On the other hand, to feed the SVM classifier with the word-level representations including word2vec, fastText, and ELMo, we computed a tf-idf-weighted average of these word representations for each sentence. Note that the reported test accuracies are for models trained on both training and validation sets using the hyper-parameters tuned based on the validation set. 



\begin{table}[t!]
    \centering
    \caption{Validation and test set accuracy of different models trained on the FarsTail training set.}
    \label{tab:overall-results}
    \begin{tabular}{P{3cm} c c c}
        \Xhline{2\arrayrulewidth}
        \textbf{Model} & \textbf{Representation} & \textbf{Val Accuracy} & \textbf{Test Accuracy}\\
        \hline
         \multirow{5}{*}{SVM} 
         & tf-idf & 0.5303 & 0.5301\\
         & LASER & 0.5459 & 0.5198\\
         & word2vec & 0.5120 & 0.5448\\
         & fastText & 0.5296 & 0.5371\\
         & ELMo & 0.5621 & 0.5710\\
         \hline
         \multirow{3}{*}{LSTM} & word2vec & 0.5172 & 0.5243\\
         & fastText & 0.5205 & 0.5192\\
         & ELMo & 0.5478 & 0.5505\\
         \hline
         \multirow{3}{*}{BiGRU} & word2vec & 0.5192 & 0.5224\\
         & fastText & 0.5211 & 0.5243\\
         & ELMo & 0.5582 & 0.5428\\
         \hline
         DecompAtt & word2vec & 0.6597 & 0.6662\\
         ESIM & fastText & 0.7033 & 0.7116\\
         HBMP & word2vec & 0.6617 & 0.6604\\
         ULMFiT & Learned & 0.7281 & 0.7244 \\
         \hline
         \multirow{2}{*}{BERT} & ParsBERT & 0.8081 & 0.8299\\
         & mBERT & \textbf{0.8263} & \textbf{0.8338}\\
         \Xhline{2\arrayrulewidth}
    \end{tabular}
\end{table}

For brevity, we just report the result of one representation for DecompAtt, ESIM, and HBMP. In the ESIM and HBMP methods, all representations obtained almost similar accuracies; while in the DecompAtt method, word2vec considerably outperformed other embeddings. According to Table~\ref{tab:overall-results}, the BERT models obtained the best accuracies with a large margin compared to other models. Between ParsBERT and mBERT, the latter shows a slightly better performance. Anyway, this 83.38\% test accuracy shows that there is a big room for improving the current methods to be useful for real-world NLP applications in different languages. 

\begin{figure}[p]
    
    \begin{subfigure}{0.53\textwidth} 
        \includegraphics[width=\textwidth]{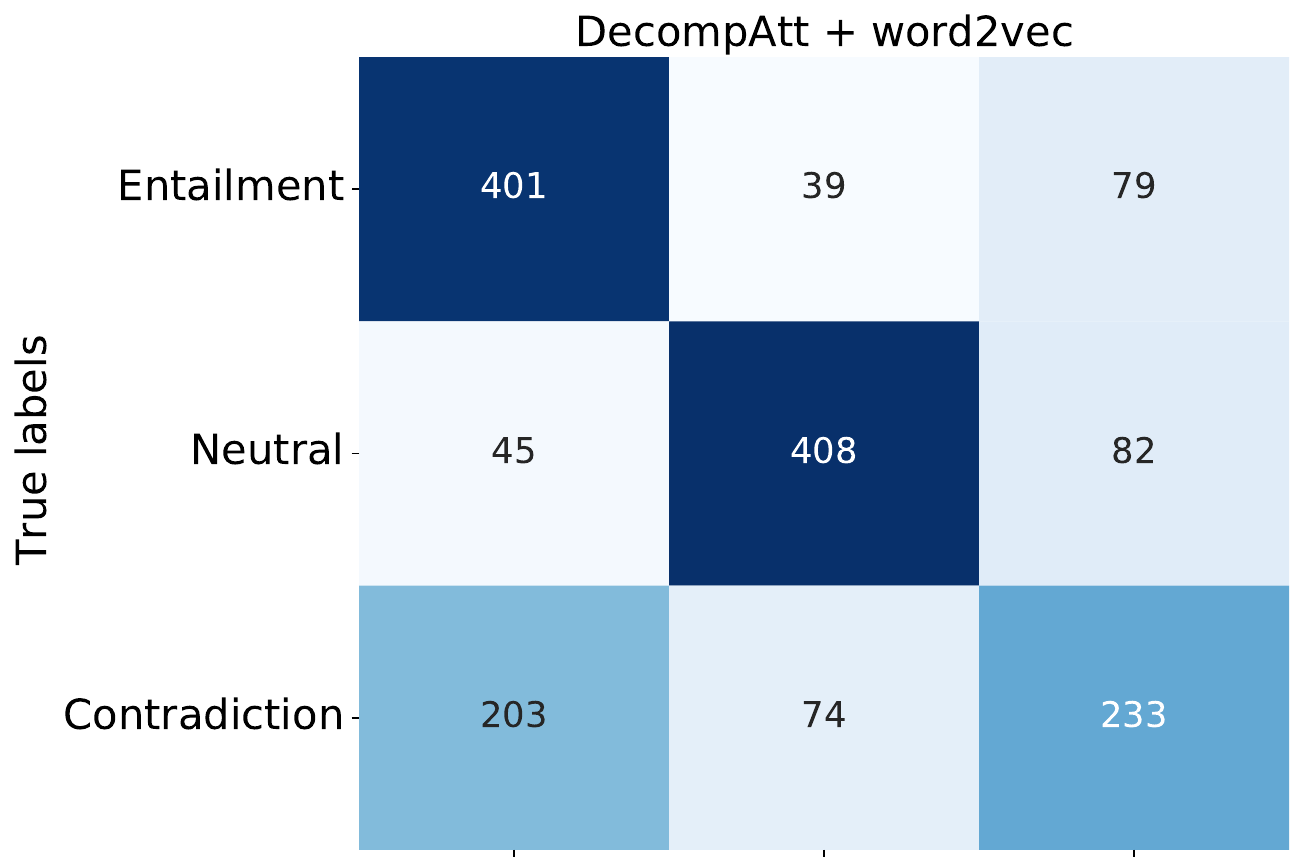}
    \end{subfigure}
    \begin{subfigure}{0.43\textwidth} 
        \includegraphics[width=\textwidth]{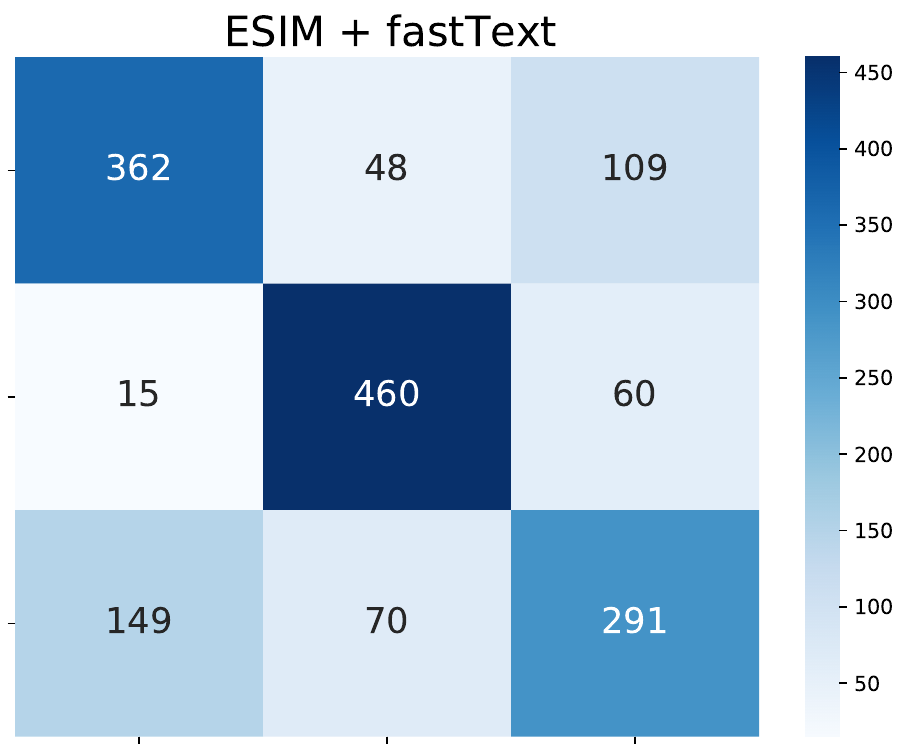}
    \end{subfigure}
    
    \begin{subfigure}{0.53\textwidth}
        \includegraphics[width=\textwidth]{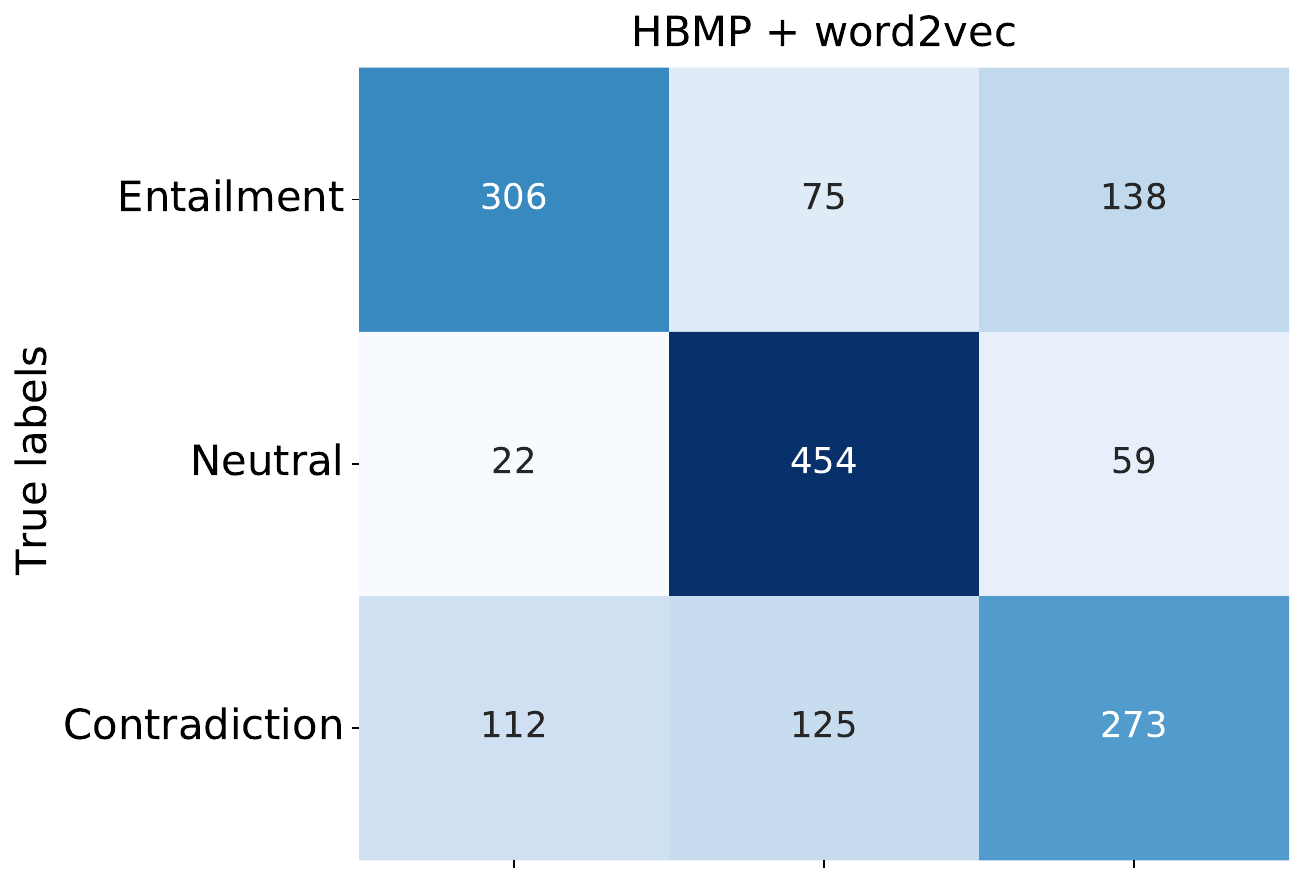}
    \end{subfigure}
    \begin{subfigure}{0.43\textwidth}
        \includegraphics[width=\textwidth]{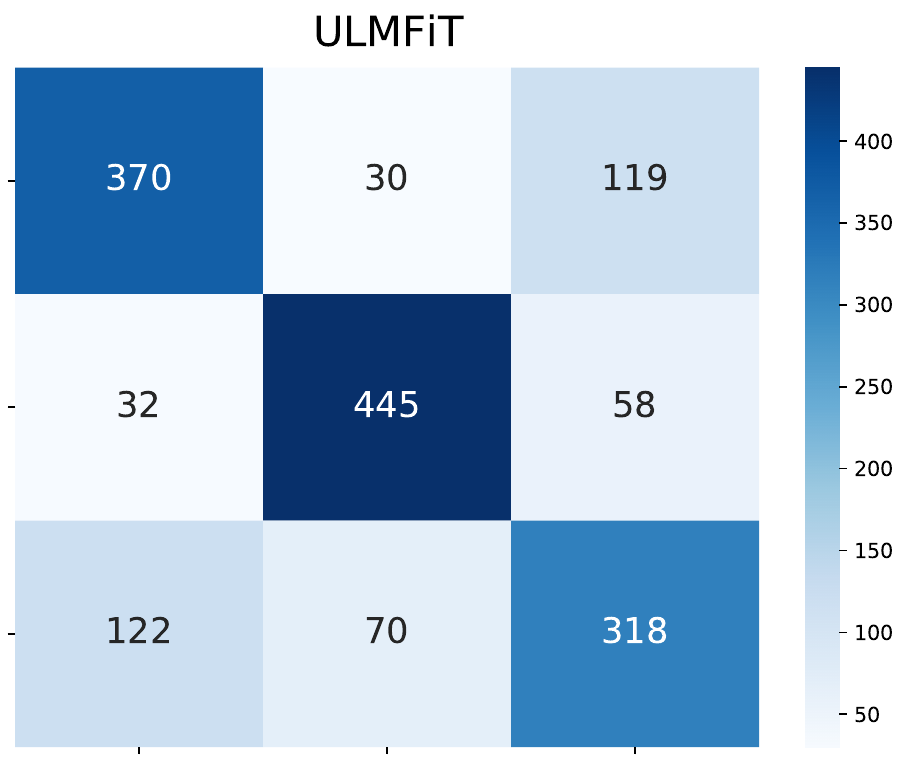}
    \end{subfigure}
    
    \begin{subfigure}{0.54\textwidth}
        \includegraphics[width=\textwidth]{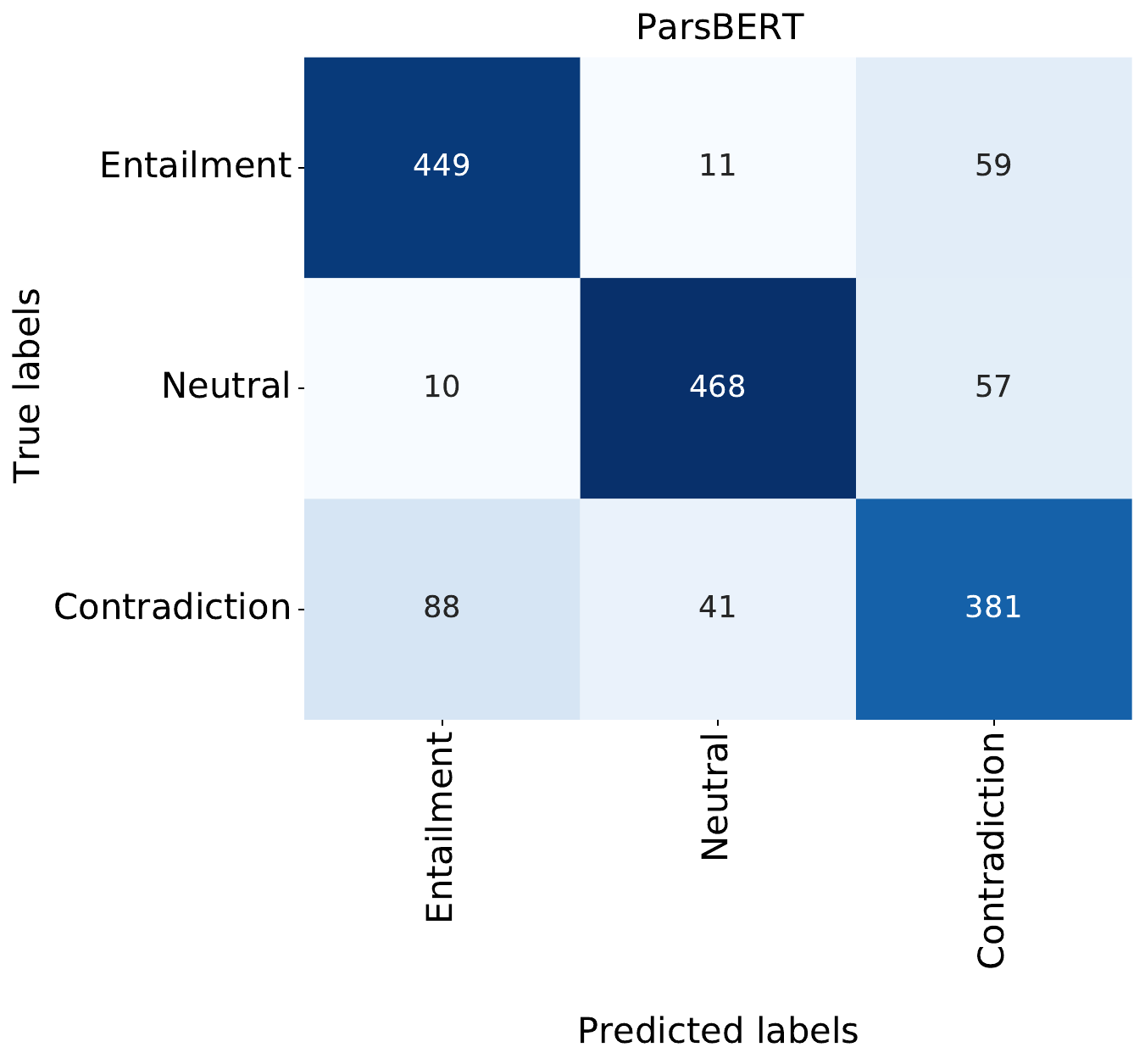}
    \end{subfigure}
    \begin{subfigure}{0.42\textwidth}
        \includegraphics[width=\textwidth]{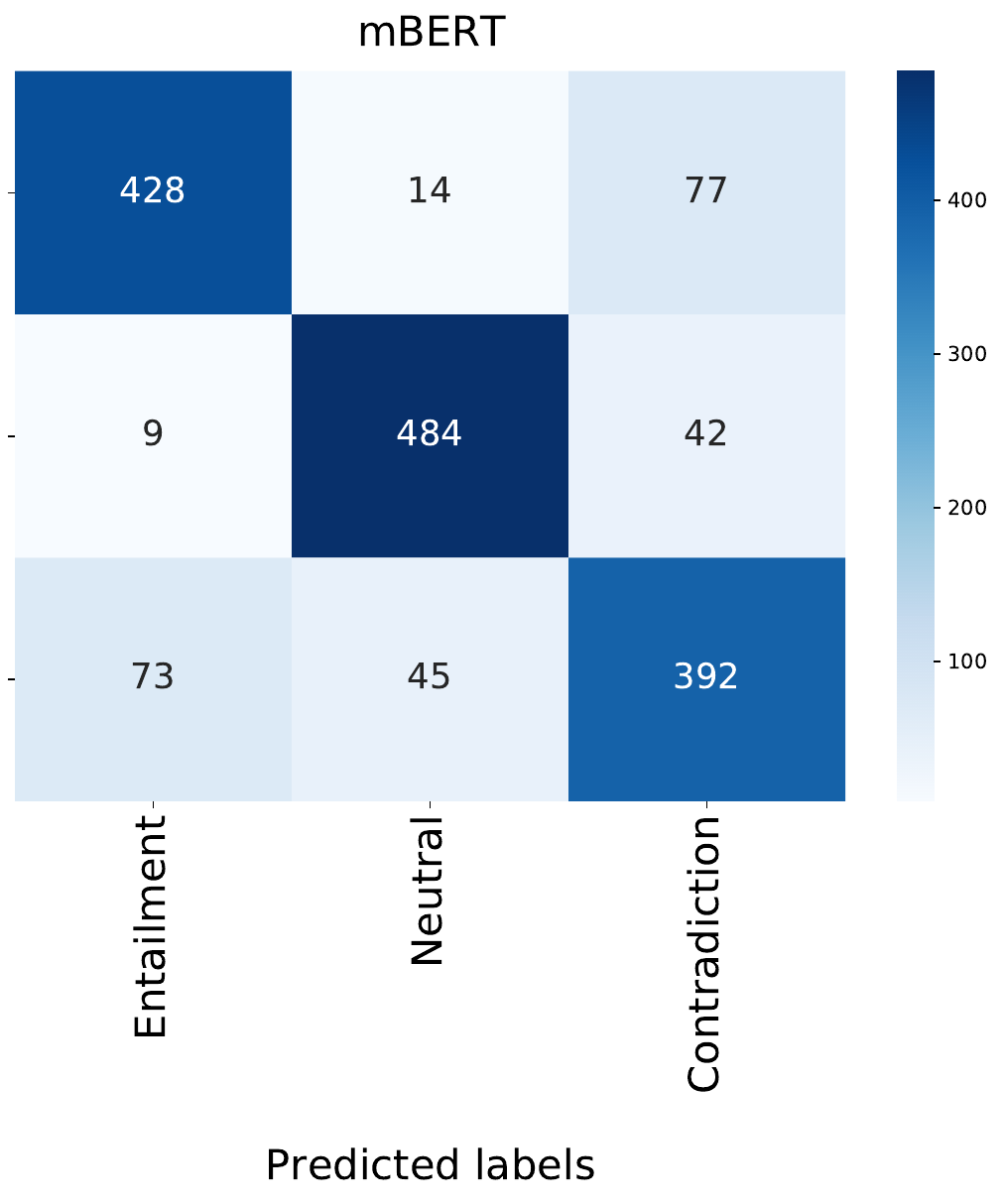}
    \end{subfigure}
    \caption{Confusion matrices of different models on the FarsTail test set.}
    \label{fig:confusion}
\end{figure}

To provide a more detailed view of the performance of different models, Fig.~\ref{fig:confusion} shows the confusion matrices of six best performing ones. According to this figure, the most difficult class for all methods is \textit{contradiction} that is confused more with \textit{entailment} than \textit{neutral}. This is because distinguishing a contradiction situation, especially from an entailment one, needs higher levels of natural language understanding than superficial pattern recognition. 

On the other hand, the \textit{neutral} class is the simplest one because many neutral samples can be easily identified by simple patterns like the overlap between their \textit{premise} and \textit{hypothesis}. This is compatible with the statistics presented in Table~\ref{tab:statistics} where the overlap between premises and hypotheses in the \textit{neutral} class is clearly different from that in the other two classes.
Obviously, the performance of the models that rely on such superficial clues can degrade in out-of-distribution situations. The next section is a step towards investigating these biases in the FarsTail dataset.

\subsection{Dataset bias}\label{subsec:bias}
Dataset bias includes correlations between input data and target values which are not generalizable to real-world instances. For example, negation words like \textit{nobody}, \textit{no}, \textit{never}, and \textit{nothing} in some NLI datasets like SNLI and MultiNLI are strongly correlated with the contradiction class~\citep{gururangan2018annotation}. Deep models tend to exploit these clues to solve the dataset instead of the intended task. Therefore, even though they obtain high in-distribution accuracies, their performance drops significantly for out-of-distribution data~\citep{mccoy2020right}. In this section, we investigate the available biases in the FarsTail dataset. 

To identify the words associated with different inference classes, the point-wise mutual information (PMI) is computed between each word and class in the training set hypotheses:

\begin{equation*}
    \textrm{PMI}(word,class) = \log\frac{p(word,class)}{p(word,.)p(.,class)}.
\end{equation*}
As in~\citep{gururangan2018annotation,bowman2020new}, we apply add-100 smoothing to the raw statistics. Table~\ref{tab:pmi} shows the top ten words by $\textrm{PMI}(word,class)$ in FarsTail as well as MultiNLI and SciTail for comparison. The table also reports the ratio of instances of each word belonging to the specified class. FarsTail shows lower PMI values and lower occurrence number of these superficial clues compared to the other two datasets. In addition, the top words by PMI in FarsTail belong to a wider range of classes. 

\begin{table}[p]
    \centering
    \caption{The top ten words by PMI(\textit{word},\textit{class}) in three datasets. The \textit{Counts} column shows how many of the instances of each word occur in hypotheses belong to the specified class.}
    \label{tab:pmi}
    \begin{tabular}{c c c c c}
        \Xhline{2\arrayrulewidth}
          & \textbf{Word} & \textbf{Class} & \textbf{PMI} & \textbf{Counts}\\
        \hline
        \multirow{10}{*}{\textbf{MultiNLI}} & never & Contradiction & 0.852 & 6599/8363\\
          & no & Contradiction&0.820 &12499/16515\\
          & nothing&Contradiction &0.775 &2090/2758\\
          & any&Contradiction &0.735 &5430/7739\\
          & none&Contradiction &0.681 &553/702\\
          & anything & Contradiction & 0.668 & 2239/3336\\
          & completely & Contradiction & 0.664 & 855/1190\\
          & also & Neutral & 0.644 & 1845/2726\\
          & refused & Contradiction & 0.644 & 401/498\\
          & nobody & Contradiction & 0.603 & 612/881\\
        \hline
        \multirow{10}{*}{\textbf{SciTail}} & to & Neutral & 0.488 & 3541/5266\\
        & have & Neutral & 0.481 & 845/1155\\
        & the & Neutral & 0.479 & 14194/21758\\
        & definite & Entailment & 0.478 & 144/146 \\
        & because & Neutral & 0.466 & 571/749\\
        & system & Neutral & 0.461 & 654/885\\
        & . & Neutral & 0.454 & 14790/23261\\
        & a & Neutral & 0.451 & 6086/9514\\
        & off & Neutral & 0.437 & 7644/12162\\
        & and & Neutral & 0.430 & 2771/4352\\
        \hline
        \multirow{10}{*}{\textbf{FarsTail}} & : & Neutral & 0.244 & 95/158\\
        & \texttt{"} & Entailment & 0.227 & 466/1053\\
        & \texttt{"} & Contradiction & 0.222 & 463/1053\\
        & \FR{تنها} (only) &Contradiction  &0.221 &61/87 \\
        & \FR{باشد} (be) &Contradiction &0.202 &202/440 \\
        & \FR{نیز} (also)&Neutral &0.179 &50/76 \\
        & \FR{فقط} (only)&Contradiction &0.168 &38/50 \\
        & \FR{خود} (self) &Neutral &0.163 &143/319 \\
        & \FR{بعد} (after)&Contradiction &0.162 &74/144 \\
        & \FR{اثر} (work,effect) &Entailment &0.159 &70/135 \\
        \Xhline{2\arrayrulewidth}
        \Xhline{2\arrayrulewidth}
    \end{tabular}
\end{table}

Even though we tried to keep the annotation clues low by reducing the amount of task-specific human-generated texts, some of these biases emerged in FarsTail hypotheses. For example, the words ``\FR{تنها}" and ``\FR{فقط}" (only) have been used to confine the general point presented in the premise to make a contradicting hypothesis as in the following instance:
\begin{center}
\includegraphics[width=0.97\textwidth]{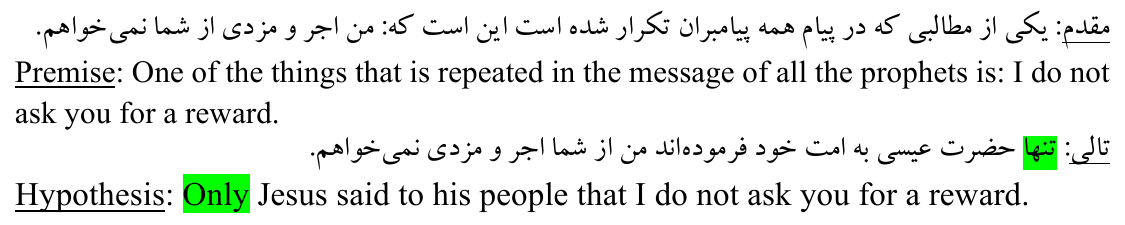}
\end{center}

\begin{figure}
    \centering
    \includegraphics[width=0.54\textwidth]{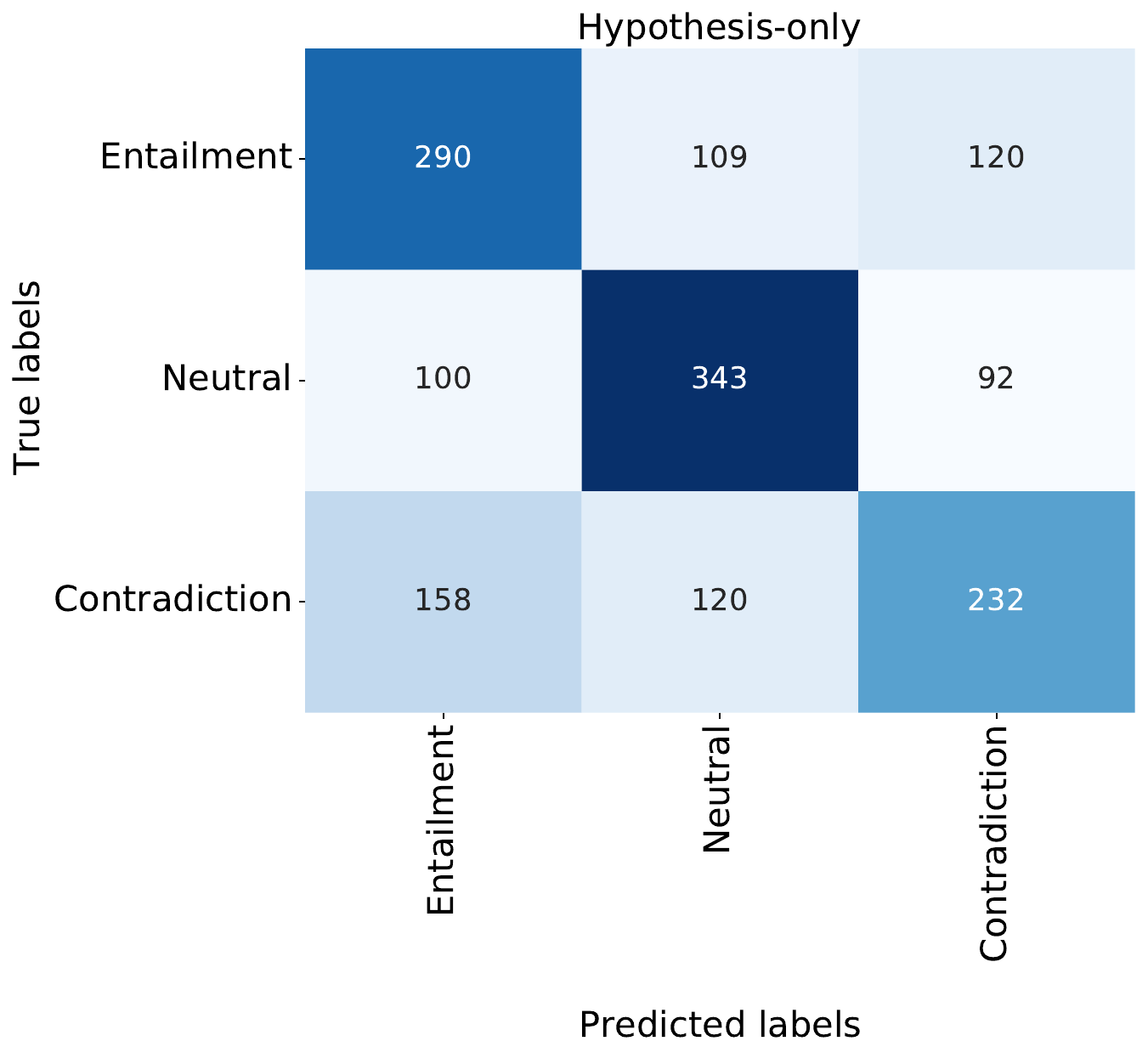}
    \includegraphics[width=0.42\textwidth]{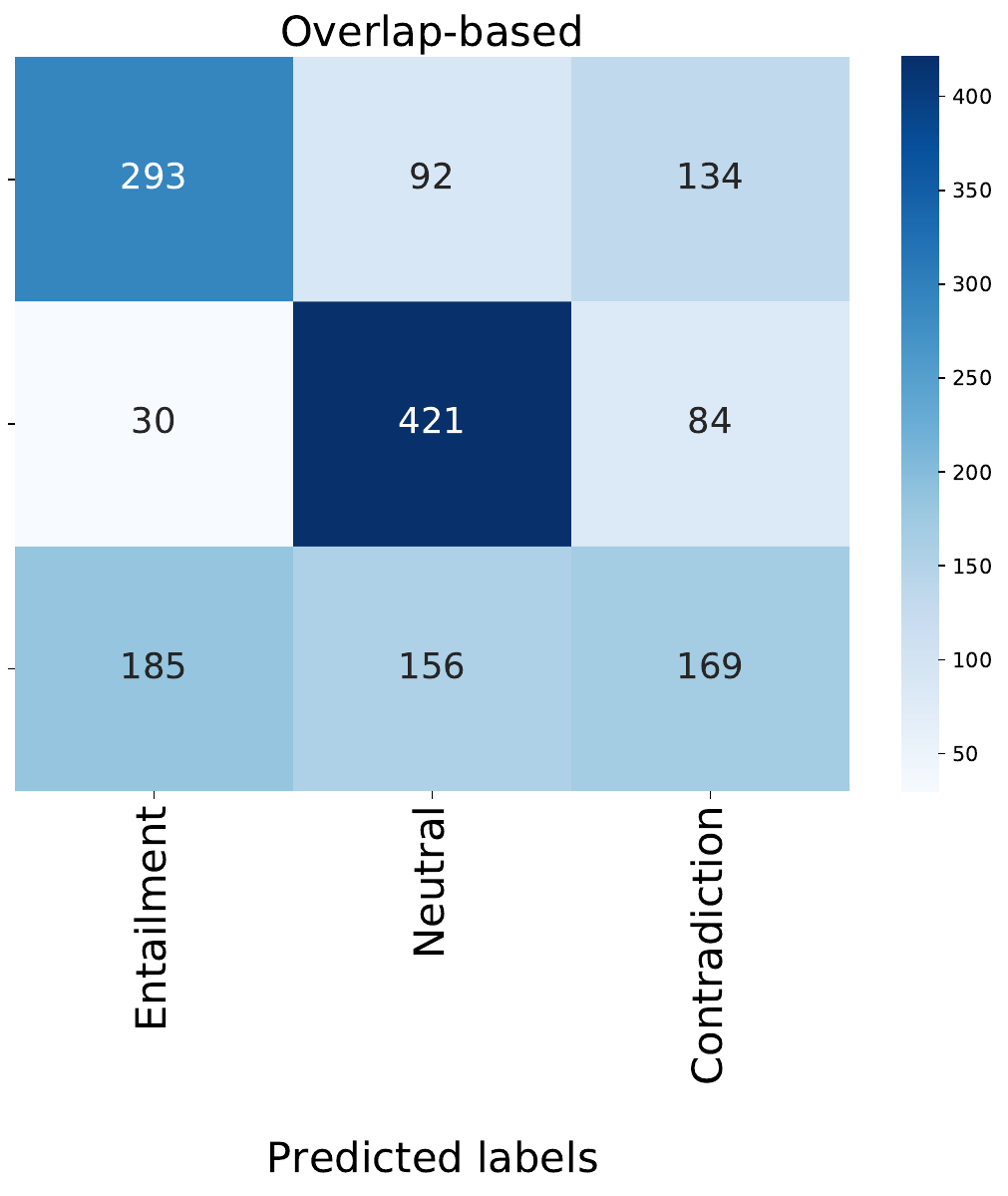}
    \caption{Confusion matrices of the biased models on the FarsTail test set.}
    \label{fig:biased-confusion}
\end{figure}

As another approach for investigating dataset biases, we evaluated two biased models which classified instances based on incomplete input data. The classification accuracy of these models gives an estimate of the degree to which the superficial clues can be exploited by the learning algorithms. Inspired from~\citep{gururangan2018annotation,poliak2018hypothesis}, we first investigated a hypothesis-only model by fine-tuning the mBERT model on the hypotheses to predict the entailment labels without seeing the premises. The model obtained an accuracy of 55.31\% on the test set. The corresponding confusion matrix presented in Fig.~\ref{fig:biased-confusion} shows that the main success of the hypothesis-only model has been in the \textit{neutral} class, with the \textit{entailment} and \textit{contradiction} classes in the next places. 


In the second biased model, we used the cosine similarity between the bag-of-word count vectors of the premise and hypothesis as the input feature to investigate the ability of a model in deciding about the inference relationship just exploiting the similarity between the premise and hypothesis. An SVM classifier trained on this input feature obtained an accuracy of 56.46\% on the test set. Fig.~\ref{fig:biased-confusion} shows that this model has obtained a good performance in distinguishing the \textit{neutral} class from the other two classes. This is compatible with the overlap statistics presented in Table~\ref{tab:statistics} where the overlap between premises and hypotheses in the \textit{neutral} class is clearly different from that in the other two classes. On the other hand, the worst performance of this biased model has been in the \textit{contradiction} class where the model has performed near random. This is because contradiction needs a higher level of inference to be determined.

According to whether or not the test samples were correctly classified by each of the biased models, we partitioned the FarsTail test set into two subsets (for each biased model): \textit{easy} and \textit{hard}. Two binary columns added to the test set, denoted as \textit{hard(hypothesis)} and \textit{hard(overlap)}, indicate whether or not each sample belongs to the \textit{hard} subset based on the \textit{hypothesis-only} and \textit{overlap-based} biased models, respectively. Comparing these subsets, 497 (32\%) test samples are easy for both biased models, while 313 (20\%) samples are hard for both. On the other hand, 386 (25\%) and 368 (23\%) test samples are hard just for the hypothesis-only and overlap-based biased models, respectively. Obviously, these two models capture different biased patterns in the dataset since nearly half of the samples are easy for one model and hard for the other. 

The introduction of these subsets of the test set allows for a more precise evaluation of the developed models. As an example, Table~\ref{tab:detailed-results} shows the detailed performance of some models on different FarsTail test subsets. As expected, all models were more successful in classifying \textit{easy} samples. This shows the previously known fact that a part of the models' success in recognizing textual entailment is due to their exploitation of available biases in the dataset~\citep{gururangan2018annotation}. Also, comparing the results obtained for the subsets respective to the two biased models shows that the models' accuracy on the \textit{hard} subset obtained by the overlap-based biased model is usually lower than that of the hypothesis-only biased model. This reveals that the models exploit more of the overlap information between premises and hypotheses than the biased patterns in the hypotheses. Obviously, these models will have difficulty in classifying samples that come from a different distribution. We consider the construction of out-of-distribution challenge sets for the FarsTail dataset as a future work. 

\begin{table}[t!]
    \centering
    \caption{Accuracy of different models on different subsets of the FarsTail test set.}
    \label{tab:detailed-results}
    \begin{tabular}{P{4.2cm} | c | c c | c c}
        \Xhline{2\arrayrulewidth}
         &  & \multicolumn{2}{c|}{\textbf{Hypothesis-only}} & \multicolumn{2}{c}{\textbf{Overlap-based}}\\
        \textbf{Model} & \textbf{Full} & \textbf{Easy} & \textbf{Hard} & \textbf{Easy} & \textbf{Hard}\\
        \hline
         DecompAtt (word2vec) & 0.6662 & 0.7341 & 0.5823 & 0.7633 & 0.5404\\
         HBMP (word2vec) & 0.6604 & 0.7618 & 0.5350 & 0.7565 & 0.5360\\
         ESIM (fastText) & 0.7116 & 0.7931 & 0.6109 & 0.8120 & 0.5815\\
         mBERT & 0.8338 & 0.8763 & 0.7811 & 0.8981 & 0.7504\\
         \Xhline{2\arrayrulewidth}
    \end{tabular}
\end{table}

\section{Conclusion}
In this paper, we introduced, to the best of our knowledge, the first relatively large-scale NLI dataset for Persian language. We presented the details of the FarsTail development process, which is carefully designed to ensure the data quality. We also presented the dataset statistics as well as the results of some traditional and state-of-the-art methods on it. We also investigated the dataset biases in FarsTail.

Due to the usage of multiple-choice questions in developing the FarsTail dataset, these questions along with their corresponding premises can also be exploited in the machine reading comprehension (MRC) task. In the future, we plan to present this MRC dataset as a byproduct of FarsTail. We also consider developing Persian NLI challenge sets as a future work to establish a benchmark for evaluating the models' out-of-distribution performance. 

Since the best obtained result on the FarsTail test set, using the powerful BERT method, is 83.38\%, we hope it invokes more research on developing methods which are applicable to real-world NLP tasks in different languages, specially data-poor ones. 

\bibliographystyle{elsarticle-num} 
\bibliography{references}

\end{document}